\documentclass{article}
\usepackage{ijcai26}

\usepackage{times}
\usepackage{soul}
\usepackage{url}
\usepackage[hidelinks]{hyperref}
\usepackage[utf8]{inputenc}
\usepackage[small]{caption}
\usepackage{graphicx}
\usepackage{amsmath}
\usepackage{amsthm}
\usepackage{amssymb}
\usepackage{booktabs}
\usepackage{algorithm}
\usepackage{algorithmic}
\usepackage{booktabs}
\usepackage{multirow}
\usepackage[table,xcdraw]{xcolor}
\usepackage[switch]{lineno}

\title{OmniOVCD: Streamlining Open-Vocabulary Change Detection with SAM 3}

\author{
Xu Zhang$^1$
\and
Danyang Li$^{2}$
\and
Yingjie Xia$^{2}$
\and
Xiaohang Dong$^1$\\
Hualong Yu$^{1}$
\and
Jianye Wang$^1$
\and
Qicheng Li$^1$\thanks{Corresponding Author}\\
$^1$TMCC, Computer Science, Nankai University ~~~~ $^2$VCIP, Computer Science, Nankai University ~~~~ \\
\textit{\{xu\_zhang, danyang.li\}@mail.nankai.edu.cn, liqicheng@nankai.edu.cn}}

\begin{document}

\maketitle

\begin{abstract}

Change Detection (CD) is a fundamental task in remote sensing. It monitors the evolution of land cover over time. Based on this, Open-Vocabulary Change Detection (OVCD) introduces a new requirement. It aims to reduce the reliance on predefined categories. Existing training-free OVCD methods mostly use CLIP to identify categories. These methods also need extra models like DINO to extract features. However, combining different models often causes problems in matching features and makes the system unstable. Recently, the Segment Anything Model 3 (SAM 3) is introduced. It integrates segmentation and identification capabilities within one promptable model, which offers new possibilities for the OVCD task. In this paper, we propose OmniOVCD, a standalone framework designed for OVCD. By leveraging the decoupled output heads of SAM 3, we propose a Synergistic Fusion to Instance Decoupling (SFID) strategy. SFID first fuses the semantic, instance, and presence outputs of SAM 3 to construct land-cover masks, and then decomposes them into individual instance masks for change comparison. This design preserves high accuracy in category recognition and maintains instance-level consistency across images. As a result, the model can generate accurate change masks. Experiments on four public benchmarks (LEVIR-CD, WHU-CD, S2Looking, and SECOND) demonstrate SOTA performance, achieving IoU scores of 67.2, 66.5, 24.5, and 27.1 (class-average), respectively, surpassing all previous methods. The code is available at \url{https://github.com/Erxucomeon/OmniOVCD}.
\end{abstract}

\section{Introduction}
\begin{figure}[t]
  \centering
   \includegraphics[width=0.85\linewidth]{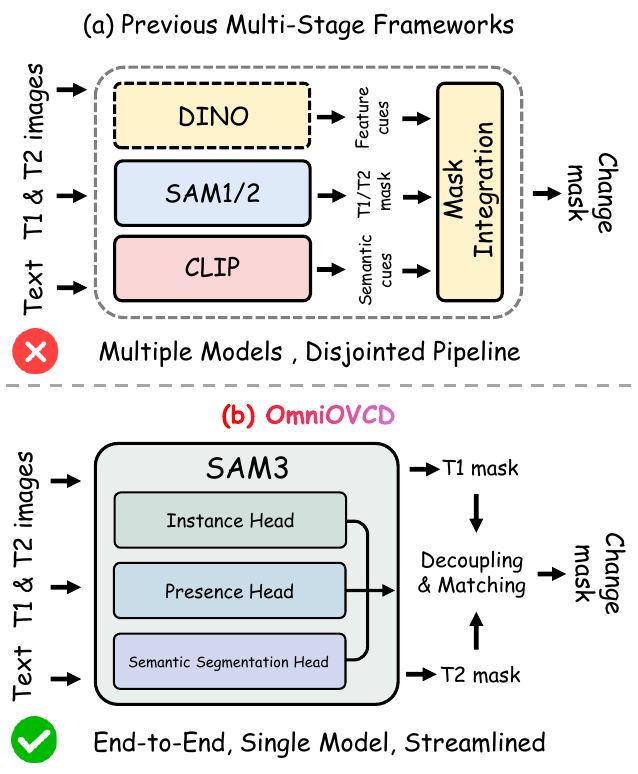}
   \caption{Architecture comparison.}
   \label{fig:architecture}
\end{figure}

Change Detection (CD) aims to accurately capture the evolution of land cover, which serves as a fundamental task for Earth observation. It supports a wide range of applications, including urban planning and land resource management~\cite{wellmann2020remote,asadzadeh2022uav,zheng2021building}. However, conventional CD methods typically operate in a closed-set setting, recognizing only the predefined categories present in the training data. This limitation prevents the models from detecting changes that were not defined during training. To address this issue, Open-Vocabulary Change Detection (OVCD) has been proposed, which leverages language guidance to detect changes of different categories, making it more suitable for open-world scenarios.

Under the supervised learning paradigm, deep neural networks have established the dominant standard for change detection~\cite{daudt2018fully,zhang2020deeply}. These methods typically leverage Siamese architectures to capture temporal variations~\cite{chen2022remote,bandara2022transformer}. Although recent methods leveraging feature interaction and fusion have achieved significant improvements~\cite{fang2023changer,xu2024hybrid}, these methods still rely heavily on the use of large-scale and high-quality annotated datasets. Manual annotation is costly; meanwhile, generalizing models to unseen categories remains difficult. Both issues hinder practical application in the open world.

\begin{figure}[t]
  \centering
   \includegraphics[width=0.9\linewidth]{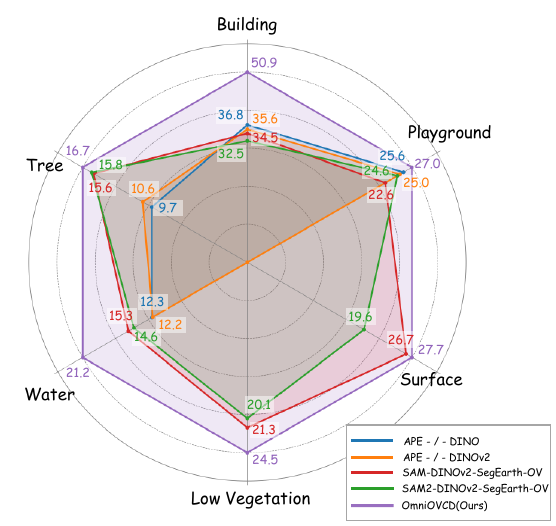}
   \caption{Compared with existing methods, OmniOVCD achieves superior IoU scores across multiple category benchmarks.}
   \label{fig:radar}
\end{figure}

The rapid development of Vision Foundation Models (VFMs), such as CLIP~\cite{radford2021learning} and SAM~\cite{kirillov2023segment,ravi2024sam,carion2025sam}, has significantly facilitated the progress of zero-shot change detection. Early pioneers such as AnyChange~\cite{zheng2024segment} and SCM~\cite{tan2024segment} explored zero-shot change detection by leveraging pre-trained foundation models for bi-temporal analysis. AnyChange extends SAM’s zero-shot segmentation to change detection via bi-temporal latent matching. It effectively locates changed regions but produces only class-agnostic masks. SCM combines SAM with CLIP, using a Piecewise Semantic Attention (PSA) scheme to provide semantic guidance and identify building-related changes. Along this line, DynamicEarth~\cite{li2025dynamicearth} formally defines the task of Open-Vocabulary Change Detection (OVCD), aiming to detect any category of interest through natural language guidance. It proposes two training-free and universal frameworks: Mask-Compare-Identify (M-C-I) and Identify-Mask-Compare (I-M-C). Both integrate various pre-trained foundation models for inference. 

While these methods have demonstrated promising performance, they still suffer from several limitations. As illustrated in Fig.~\ref{fig:architecture}(a), these frameworks rely on combining independent models, which often creates additional computational cost and compromises system stability. Specifically, these multi-step pipelines face difficulties in aligning features across modules. Because each component operates independently, errors accumulate across stages, ultimately reducing overall accuracy.  The IoU performance of the four best-performing configurations from DynamicEarth~\protect\cite{li2025dynamicearth} can be seen in Fig.~\ref{fig:radar}. The ``Building'' score represents the average IoU of the building category across all datasets, and the remaining categories are evaluated on the SECOND dataset. These configurations achieve competitive overall IoU. However, each configuration exhibits noticeable variation across different categories. This variation demonstrates the instability and limited generalization of existing OVCD pipelines.

To address the instability and complexity of existing models, we introduce OmniOVCD. As illustrated in Fig.~\ref{fig:architecture}(b), OmniOVCD is a standalone framework that leverages the unified architectural advancements of the Segment Anything Model 3 (SAM 3). It simplifies the architecture and streamlines the open-vocabulary change detection process. The core of our framework lies in the Synergistic Fusion to Instance Decoupling (SFID) strategy. SFID performs synergistic fusion by combining semantic, instance, and presence outputs. It then separates them into instance masks to ensure accurate instance matching during bi-temporal comparison. As shown in Fig.~\ref{fig:radar}, this strategy effectively reduces pseudo-changes, enabling OmniOVCD to achieve consistently strong performance across different categories. Overall, our contributions are as follows:
\begin{itemize}
\item[$\bullet$] We propose OmniOVCD, the first standalone framework for open-vocabulary change detection based on SAM 3. By using the concept prompts in SAM 3, we successfully streamline the change detection process, achieving a simpler and more efficient workflow.
\item[$\bullet$] We introduce the SFID strategy, which first fuses the multi-head outputs and then separates them into instance masks. This design enables high-quality instance-level reasoning, which significantly improves the accuracy of change detection.
\item[$\bullet$] Comprehensive experiments show that OmniOVCD achieves state-of-the-art performance, outperforming existing training-free methods on both building and multi-class benchmarks.
\end{itemize}

\section{Related Work}
\subsection{Segment Anything Model} 
The Segment Anything Model (SAM)~\cite{kirillov2023segment} establishes a promptable paradigm for zero-shot image segmentation. Efficient variants such as FastSAM~\cite{zhao2023fast} and MobileSAM~\cite{mobile_sam} address the high inference costs of the original architecture. These models enable real-time performance on resource-constrained platforms. Subsequently, SAM 2~\cite{ravi2024sam} extended the promptable framework to the temporal dimension. It incorporates a memory attention mechanism for video object tracking and mask propagation. Recently, SAM 3~\cite{carion2025sam} unifies detection, segmentation, and tracking within a single architecture through Promptable Concept Segmentation (PCS). This approach generates high-quality masks and provides unique identities for matching instances. While SAM and its derivatives focus on single-image segmentation, their powerful zero-shot capabilities create new possibilities for change detection. AnyChange~\cite{zheng2024segment} and SCM~\cite{tan2024segment} established early training-free paradigms by utilizing the foundational knowledge of SAM and FastSAM. Specifically, AnyChange relies on bi-temporal latent matching to locate changed regions, but it only produces class-agnostic masks. In contrast, SCM employs a Piecewise Semantic Attention (PSA) scheme to identify building-specific changes. The promptable concept segmentation capability of SAM 3 establishes a solid foundation for OmniOVCD. It simplifies the alignment issues typical of multi-model change detection.
\begin{figure*}[t]
  \centering
   \includegraphics[width=1.0\linewidth]{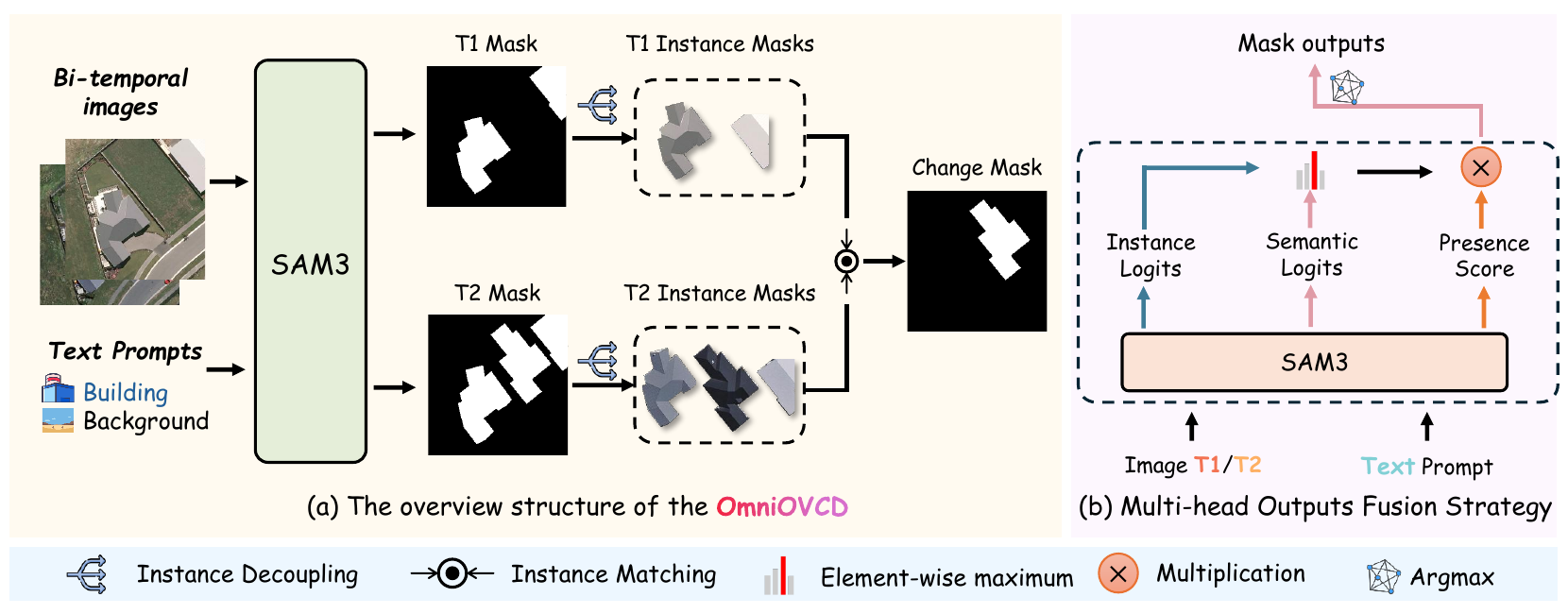}
   \caption{(a) shows the overview of OmniOVCD framework. The model takes bi-temporal images and corresponding text prompts to generate initial masks via SAM 3. These masks are then separated into instance masks for instance-level comparison, which produce the accurate change detection mask. (b) shows the multi-head outputs fusion strategy. This strategy fuses the semantic and instance head outputs from SAM 3 and uses the presence head outputs for filtering. This approach effectively improves the accuracy of single-image segmentation.}
   \label{fig:overview}
\end{figure*}

\subsection{Traditional Change Detection}
The change detection task has traditionally been partitioned into two sub-tasks: binary change detection and semantic change detection. Early change detection methods are based on siamese convolutional neural networks, such as FC-Siam-diff \cite{daudt2018fully} and IFN \cite{zhang2020deeply}. These models use weight-sharing encoders to compare bi-temporal images. Transformer-based architectures, such as BiT \cite{chen2022remote}, ChangeFormer \cite{bandara2022transformer}, and Changer \cite{fang2023changer}, improve performance by capturing global context. These models show better feature interaction and discriminative power on various land-cover benchmarks \cite{feng2023change,xu2024hybrid}. Recently, researchers have applied vision foundation models to change detection. SAM-CD \cite{ding2024adapting} and SCD-SAM \cite{mei2024scd} leverage pre-trained features to enhance change analysis \cite{peng2021scdnet,ding2022bi}. However, these methods rely on predefined categories and cannot recognize changes outside their training data. This limitation leads to poor performance when they face complex scenarios.

\subsection{Open-Vocabulary Change Detection}
To better understand complex remote sensing images, recent research has shifted from identifying predefined categories to open-vocabulary analysis. Early OVSS methods such as MaskCLIP~\cite{zhou2022extract} and SCLIP~\cite{wang2024sclip} use CLIP~\cite{radford2021learning} to extract pixel-level features. Subsequently, ProxyCLIP~\cite{lan2024proxyclip} uses auxiliary structural guidance from DINO~\cite{caron2021emerging} or SAM~\cite{kirillov2023segment} to improve localization. SegEarth-OV~\cite{li2025segearthov} optimizes CLIP features to better represent the diverse shapes and sizes of objects in remote sensing images. SegEarth-OV3~\cite{li2025segearthov3} fuses the multi-head outputs of SAM 3 to improve zero-shot recognition. Following these works, DynamicEarth~\cite{li2025dynamicearth} formally defines the task of open-vocabulary change detection (OVCD). This work introduces two training-free frameworks: Mask-Compare-Identify (M-C-I) and Identify-Mask-Compare (I-M-C). However, these methods rely on combining different pre-trained models, which leads to high complexity and instability. Experiments in DynamicEarth reveal that such a framework lacks consistency, as its performance varies significantly across datasets. This is because these separate components are not well-aligned, leading to accumulated errors during inference.

\section{Methods}
\subsection{Task Definition}
Open-vocabulary change detection (OVCD) identifies changes between two images based on a text description. This task requires both locating changed regions and identifying their categories. Formally, the input consists of two remote sensing images $x_{img}^1, x_{img}^2 \in R^{H \times W \times C}$. These images cover the same area and are captured at different times $T_1$ and $T_2$. Unlike traditional change detection paradigms that are restricted to a predefined set of categories, the OVCD task introduces a flexible text prompt, $x_{txt}$, which provides natural language guidance for the target categories of interest. This task requires the model to work outside fixed categories and identify new types of changes in diverse real-world environments. The primary objective is to generate a high-fidelity change mask $M$. This mask must locate where changes occur and identify what categories they represent, as specified by the text $x_{txt}$. This inference process can be generally formulated as:
\begin{equation}
M = \Phi_{OmniOVCD}(x_{img}^1, x_{img}^2, x_{txt}),
\end{equation}
where $\Phi_{OmniOVCD}$ denotes the proposed standalone architecture, as illustrated in Fig.~\ref{fig:overview}(a). It accepts bi-temporal images along with their corresponding text prompts to generate initial semantic masks using SAM 3. These masks are subsequently split into individual instance masks, which are compared across time to produce the final change mask.
\subsection{SFID Pipeline}
The Synergistic Fusion to Instance Decoupling (SFID) pipeline enables instance-level change analysis from dense semantic predictions. This process consists of two sequential phases: \textbf{Synergistic Mask Fusion} and \textbf{Instance Decoupling and Matching}. This strategy adopts the synergistic perception paradigm established in SegEarth-OV3~\cite{li2025segearthov3} and extends its application to a bi-temporal context for dual-image processing.  Initially, the framework fuses SAM 3's semantic, instance, and presence outputs from each image to construct high-fidelity semantic maps. To convert pixel-wise class predictions into discrete object entities, these continuous semantic regions are decoupled into individual instance masks based on topological connectivity. Subsequently, a bi-temporal consistency analysis evaluates each instance by measuring how much masks overlap at the same geographic location. Instances that stay in the same place are marked as unchanged, while those without a matching counterpart are identified as valid change candidates. 

\textbf{Synergistic Mask Fusion.} To establish a precise representation of the land-cover in both images, the framework adopts a synergistic perception strategy that fuses the multi-head outputs of SAM 3, as shown in Fig.~\ref{fig:overview}(b). This approach is specifically designed to handle the structural duality inherent in remote sensing imagery. In these scenes, discrete objects and expansive land-cover regions are both present and require different types of analysis. Specifically, discrete entities such as buildings require precise boundary localization at the instance level. In contrast, land-cover regions like water or vegetation need pixel-wise semantic continuity for an accurate representation. The framework applies this synergistic mechanism to each image $x_{img}^t$ individually. This fusion ensures that the fused semantic masks maintain sharp instance boundaries while achieving high categorical accuracy. By leveraging these complementary strengths, the model generates a comprehensive land-cover description. Detailed shapes of individual objects are accurately preserved. At the same time, the broader semantic context provides consistency and coherence across the scene, allowing the model to capture both fine-grained structural details and global information.

The initial stage of this process involves the aggregation of instance-centric cues generated by the Transformer-based decoder. For each temporal branch $t$, the model produces a set of $N_t$ discrete instance queries, denoted as $\mathcal{P}_{inst}^t = \{(P_{inst, t}^{(k)}, s_{conf, t}^{(k)})\}_{k=1}^{N_t}$. $P_{inst, t}^{(k)} \in [0, 1]^{H \times W}$ represents the spatial probability map for the $k$-th instance and $s_{conf, t}^{(k)}\in [0, 1]$ signifies its confidence score. To transform these sparse instance-level predictions into a category-level representation, a weighted maximal selection strategy is employed. The aggregated instance map at time $t$, denoted as $P_{agg}^t$, is formulated by computing the peak weighted response across all queries for each pixel location $(h, w)$:
\begin{equation}
P_{agg}^t(h,w) = \max_{k=1}^{N_t} \left( P_{inst, t}^{(k)}(h,w) \cdot s_{conf, t}^{(k)} \right).
\end{equation}

In parallel with instance aggregation, the semantic segmentation head generates a dense probability map $P_{sem}^t \in [0, 1]^{H \times W}$ for each image. This map helps preserve the spatial integrity of continuous land-cover categories. While $P_{agg}^t$ is effective at capturing the geometric boundaries of individual targets such as buildings, the instance-level representation can produce fragmented or discontinuous masks in large, amorphous regions. In contrast, $P_{sem}^t$ offers complete coverage for categories like water or low vegetation, but may exhibit boundary adherence or semantic blurring in densely populated object areas. To leverage their complementary structural properties, a pixel-wise max-fusion operation is applied. This operation is performed across both temporal states to obtain an intermediate fused representation $P_{fused, t}$:
\begin{equation}
P_{fused, t}(h,w) = \max \left( P_{sem, t}(h,w), P_{agg, t}(h,w) \right).
\end{equation}

Furthermore, to reduce semantic noise in open-vocabulary inference, the framework employs a bi-temporal presence gating mechanism to handle sparse category activation. Local remote sensing image patches usually contain only a small subset of categories from the global vocabulary $\mathcal{C}$.
As a result, the model may incorrectly predict land-cover types that are not actually present, especially when visual patterns are similar across categories. For each queried concept $c$, the presence head of SAM 3 provides a global existence score $S_{pres, t} \in [0, 1]$, indicating the probability that the category exists within the image $x_{img}^t$. This score is utilized as a soft gating factor to recalibrate the fused probability maps. The final output $M_t$ is obtained in a pixel-wise manner. Each pixel is assigned to the category with the highest probability after gating:
\begin{equation}
\begin{split}
P_{final, t}^{(c)} &= P_{fused, t}^{(c)} \cdot S_{pres, t}^{(c)}, \\
M_t(h,w) &= \arg \max_{c \in \mathcal{C}} P_{final, t}^{(c)}(h,w).
\end{split}
\end{equation}

 This fusion and gating pipeline produces clean and consistent semantic maps. These representations provide a reliable basis for detecting accurate changes in subsequent stages.
 
\begin{table*}[t]
\centering
\setlength{\tabcolsep}{12pt}
\begin{tabular}{c|cc|cc|cc}
\toprule
                         & \multicolumn{2}{c|}{LEVIR-CD} & \multicolumn{2}{c|}{WHU-CD} & \multicolumn{2}{c}{S2Looking} \\ \cmidrule(l){2-7} 
\multirow{-2}{*}{Method} & IoU           & F1           & IoU          & F1          & IoU           & F1            \\ \midrule
PCA-KM~\cite{celik2009unsupervised}                   & 4.8           & 9.1          & 5.4          & 10.2        & -             & -             \\
CNN-CD~\cite{el2016convolutional}                   & 7.0           & 13.1         & 4.9          & 9.4         & -             & -             \\
DSFA~\cite{du2019unsupervised}                     & 4.3           & 8.2          & 4.1          & 7.8         & -             & -             \\
DCVA~\cite{saha2019unsupervised}                     & 7.6           & 14.1         & 10.9         & 19.6        & -             & -             \\
GMCD~\cite{tang2021unsupervised}                     & 6.1           & 11.6         & 10.9         & 19.7        & -             & -             \\
CVA~\cite{bovolo2006theoretical}                      & -             & 12.2         & -            & -           & -             & 5.8           \\
DINOv2+CVA~\cite{zheng2024segment}               & -             & 17.3         & -            & -           & -             & 4.3           \\
AnyChange-H~\cite{zheng2024segment}              & -             & 23.0         & -            & -           & -             & 6.4           \\
SCM~\cite{tan2024segment}                      & 18.8          & 31.7         & 18.6         & 31.3        & -             & -             \\ 
*SAM-DINO-SegEarth-OV~\cite{li2025dynamicearth}    & 33.0          & 49.7         & 35.8         & 52.8        & 22.5          & 36.7          \\
*SAM-DINOv2-SegEarth-OV~\cite{li2025dynamicearth}  & 36.6          & 53.6         & 38.8         & 55.9        & 23.9          & 38.5          \\
*SAM2-DINOv2-SegEarth-OV~\cite{li2025dynamicearth} & 33.8          & 50.5         & 36.8         & 53.8        & 23.1          & 37.6          \\
*APE - / - DINO~\cite{li2025dynamicearth}          & 53.5          & 69.7         & 54.5         & 70.5        & 10.1          & 18.4          \\
*APE - / - DINOv2~\cite{li2025dynamicearth}        & 50.0          & 66.7         & 55.2         & 71.1        & 5.3           & 10.1          \\ \rowcolor{blue!5}
OmniOVCD                  & \textbf{67.2}          & \textbf{80.4}         & \textbf{66.5}         & \textbf{79.9}        & \textbf{24.5}          & \textbf{39.4}          \\ \bottomrule
\end{tabular}
\caption{Comparison on building change detection datasets. ``*'' denotes re-implementation under identical settings.}
\label{tab:building}
\end{table*}

\textbf{Instance Decoupling and Matching.} The previous stage produces clean semantic probability maps. However, these maps represent categories at the pixel level and do not distinguish individual instances.
 To capture geospatial transitions at the instance level, the framework applies an instance decoupling mechanism. This mechanism divides the continuous semantic map into separate regions based on topological connectivity. Each region corresponds to a separate instance. Specifically, for each binary semantic mask $M_t \in \{0, 1\}^{H \times W}$ ($t \in \{1, 2\}$), the framework performs connected-component analysis using an 8-connectivity rule. This process extracts a set of ${K_t}$ candidate instances, denoted as $\mathcal{I}_t = \{I_{t, i} \mid \Psi(M_t) \}_{i=1}^{K_t}$. $\Psi(\cdot)$ denotes the operation that extracts connected regions from the mask. 

Crucially, the instances extracted at this stage are not limited to discrete categories like buildings; they also include continuous land-cover types such as water or low vegetation. Raw instance-level predictions from foundation models often produce fragmented masks in continuous regions. Our synergistic mask fusion stage addresses this by fusing SAM 3's multi-head outputs, producing sharper and more stable boundaries. As a result, the decoupling of these refined semantic masks enables the extraction of coherent instances even for amorphous categories. It provides a robust representation for complex geospatial analysis. 

Our change detection operates at the instance level. It compares objects across time rather than relying on pixel-wise differences. For each instance $I_{1,i} \in \mathcal{I}_1$, we measure its spatial overlap with all instances in $\mathcal{I}_2$. We define the overlap ratio between two instances as:
\begin{equation}
\mathcal{R}(I_{t,a}, I_{t',b}) = |I_{t,a} \cap I_{t',b}|/|I_{t,a}|,
\end{equation}
which represents the fraction of $I_{t,a}$ covered by $I_{t',b}$. During the forward check from $T_1$ to $T_2$, an instance in $\mathcal{I}_1$ is considered unchanged if it has at least one match in $\mathcal{I}_2$ with $\mathcal{R} \geq \tau_{match}$; otherwise, it is included in the set of T1 changes, denoted as $\mathcal{C}_{T1}$. Similarly, during the backward check from $T_2$ to $T_1$, instances in $\mathcal{I}_2$ with no matching $T_1$ instance above the threshold are included in the set of T2 changes, denoted as $\mathcal{C}_{T2}$. The bi-temporal change mask $M_{change}$ is then generated by combining all instances from both sets. Formally, the change mask is defined as:
\begin{equation}
M_{change}(h, w) = \mathbb{I} \left( \exists I \in (\mathcal{C}_{T1} \cup \mathcal{C}_{T2}) \text{ s.t. } I(h, w) = 1 \right),
\end{equation}
where $\mathbb{I}(\cdot)$ is the indicator function, mapping the existential condition to the binary domain $\{0,1\}$. This instance-level matching ensures that $M_{change}$ captures accurate semantic transitions while suppressing false changes caused by lighting or seasonal variations.

\section{Experiments}
\subsection{Experimental Setting}
We adopt the Segment Anything Model 3 (SAM 3)~\cite{carion2025sam} as the foundational model for OmniOVCD. All experiments are conducted on a single NVIDIA GeForce RTX 3090 GPU with 24GB of VRAM.

\begin{table*}[t]
\centering
\setlength{\tabcolsep}{3pt}
\begin{tabular}{c|cc|cc|cc|cc|cc|cc|cc}
\toprule
 &
  \multicolumn{2}{c|}{Building} &
  \multicolumn{2}{c|}{Tree} &
  \multicolumn{2}{c|}{Water} &
  \multicolumn{2}{c|}{Low vegetation} &
  \multicolumn{2}{c|}{Surface} &
  \multicolumn{2}{c|}{Playground} &
  \multicolumn{2}{c}{Class Avg} \\ \cmidrule(l){2-15} 
\multirow{-2}{*}{Method} &
  IoU &
  F1 &
  IoU &
  F1 &
  IoU &
  F1 &
  \phantom{75}IoU &
  F1 &
  IoU &
  F1 &
  IoU &
  F1 &
  IoU &
  F1 \\ \midrule
*SAM-DINOv2-SegEarth-OV &
  38.8 &
  56.0 &
  15.6 &
  27.0 &
  15.3 &
  26.5 &
  \phantom{75}21.3 &
  35.1 &
  26.7 &
  42.1 &
  22.6 &
  36.8 &
  23.4 &
  37.3 \\
*SAM2-DINOv2-SegEarth-OV &
36.3 &
53.3 &
15.8 &
27.3 &
14.6 &
25.5 &
\phantom{75}20.1 &
33.4 &
19.6 &
32.7 &
24.6 &
39.5 &
21.8 &
35.3 \\ 
*APE - / - DINO          & 29.1 & 45.1 & 9.7  & 17.7 & 12.3 & 22.0 &\phantom{75}-    &-    &-    &-    & 25.6 & 40.8 & 12.8 & 21.0\\
*APE - / - DINOv2        & 31.9 & 48.3 & 10.6 & 19.2 & 12.2 & 21.7 & \phantom{75}-    &-    &-    &-    & 25.0 & 40.0 & 13.3 & 21.5\\  \rowcolor{blue!5}
OmniOVCD &
  \textbf{45.2} &
  \textbf{62.3} &
  \textbf{16.7} &
  \textbf{28.6} &
  \textbf{21.2} &
  \textbf{35.0} &
  \phantom{75}\textbf{24.5} &
  \textbf{39.3} &
  \textbf{27.7} &
  \textbf{43.4} &
  \textbf{27.0} &
  \textbf{42.4} &
  \textbf{27.1} &
  \textbf{41.8} \\ \bottomrule
\end{tabular}
\caption{Comparison on SECOND datasets. ``*'' denotes re-implemented results under identical settings, including the two best-performing M-C-I architectures and the two best-performing I-M-C architectures from DynamicEarth~\protect\cite{li2025dynamicearth}. ``-'' denotes that the score is close to 0.}
\label{tab:second}
\end{table*}

\subsection{Datasets}
\textbf{LEVIR-CD.} LEVIR-CD~\cite{chen2020spatial} is a high-resolution dataset for building change detection, containing 637 bi-temporal image pairs from various cities in Texas, USA. Each image has a resolution of 0.5 m/pixel and a size of $1024 \times 1024$ pixels, with a total of 31333 annotated change instances. The dataset is split into 445 training pairs, 64 validation pairs, and 128 test pairs. 

\textbf{WHU-CD.} WHU-CD~\cite{ji2019fully} is a large-scale urban change detection dataset. It is built from aerial images captured before and after a magnitude 6.3 earthquake in Christchurch, New Zealand. The dataset consists of a single bi-temporal image pair with a spatial resolution of $0.075~\text{m}$. The image size is $32507 \times 15354$ pixels and covers an area of approximately $20.5~\text{km}^2$. For experiments, the images are divided into $256 \times 256$ pixel patches. These patches are split into training, validation, and test sets with a ratio of 8:1:1.

\textbf{S2Looking.} S2Looking~\cite{shen2021s2looking} is a large-scale satellite change detection dataset collected from rural regions worldwide. It contains 5000 bi-temporal image pairs, each with a size of $1024 \times 1024$ pixels and spatial resolutions ranging from 0.5 to 0.8 meters. The images are captured from side-looking viewpoints with varying off-nadir angles, rather than standard near-nadir views. This imaging setup introduces geometric distortions and illumination variations, making change detection more challenging in rural scenes. The dataset is split into 3500 training pairs, 500 validation pairs, and 1000 test pairs.

\textbf{SECOND.} The SECOND~\cite{yang2022asymmetric} dataset is used to evaluate the model's ability to detect detailed category changes in urban and suburban areas. It contains 4662 image pairs of size $512 \times 512$ from Hangzhou, Chengdu, and Shanghai. The dataset is split into 2968 training pairs and 1694 test pairs. Following the DynamicEarth~\cite{li2025dynamicearth}, we evaluate binary change detection for each category to measure the model's accuracy across different land-cover types.

\subsection{Evaluation Metrics}
\textbf{IoU.} Intersection over Union (IoU) is a metric used to measure the overlap between predicted and reference regions. It is calculated using True Positives (TP, correctly predicted pixels), False Positives (FP, incorrectly predicted pixels), and False Negatives (FN, missed pixels). IoU is defined as the ratio of the intersection to the union of the predicted and reference regions:
\begin{equation}
    \text{IoU} = \text{TP}/(\text{TP + FP + FN}).
\end{equation}

\textbf{F1.} The F1 Score is a metric that harmonizes Precision (P) and Recall (R). Precision measures the proportion of correctly predicted positive pixels among all predicted positives, while Recall measures the proportion of correctly predicted positive pixels among all actual positives. The F1 Score is the harmonic mean of Precision and Recall, defined as:
\begin{equation}
\begin{aligned}
    \text{P} &= \text{TP / (TP + FP)}, \\
    \text{R} &= \text{TP / (TP + FN)}, \\
    \text{F1} &= \text{2} \times \text{P} \times \text{R} / (\text{P} + \text{R}). \\
\end{aligned}
\end{equation}
\subsection{Compared Methods}
To assess the performance of OmniOVCD, we select some unsupervised methods for comparison, including PCA-KM~\cite{celik2009unsupervised}, CNN-CD~\cite{el2016convolutional}, DSFA~\cite{du2019unsupervised}, DCVA~\cite{saha2019unsupervised}, GMCD~\cite{tang2021unsupervised}, CVA~\cite{bovolo2006theoretical}, AnyChange~\cite{zheng2024segment} and SCM~\cite{tan2024segment}. 
We further compare our results with the state-of-the-art OVCD method, DynamicEarth~\cite{li2025dynamicearth}. We re-implement the five best configurations of DynamicEarth on three building change detection datasets and its four best configurations on the SECOND dataset.

\begin{figure}[t]
  \centering
   \includegraphics[width=0.99\linewidth]{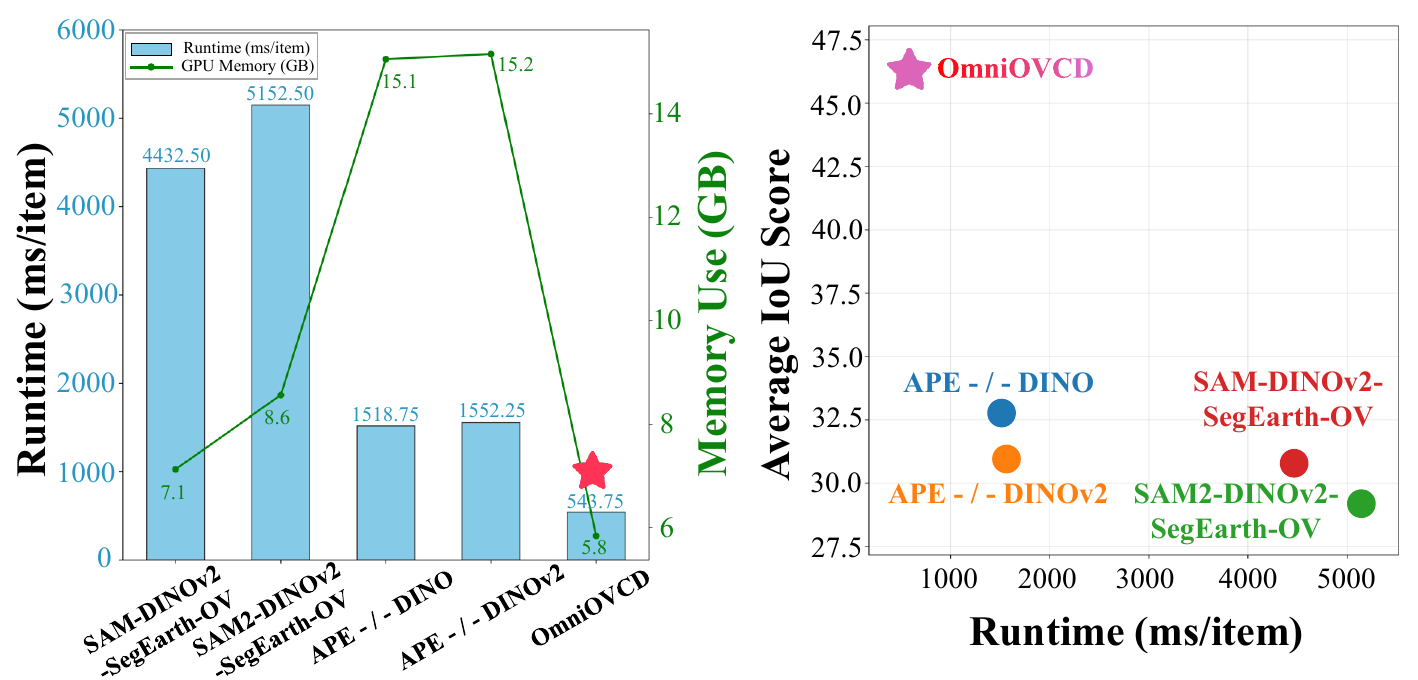}
   \caption{Comparison of GPU memory usage, inference
efficiency and IoU performance on an 3090 GPU.}
   \label{fig:cost}
\end{figure}

\subsection{Results} 
\textbf{Building Change Detection.} As shown in Tab.~\ref{tab:building}, our proposed OmniOVCD achieves SOTA performance on all three benchmarks. It consistently outperforms both traditional unsupervised methods and recent open-vocabulary models. Early methods, such as PCA-KM~\cite{celik2009unsupervised} and DSFA~\cite{du2019unsupervised}, perform poorly on these complex building change tasks, with IoU scores often below 15. DynamicEarth~\cite{li2025dynamicearth} improves results, but its performance is less stable across different datasets.

Specifically, the SAM-DINOv2-SegEarth-OV configuration achieves a relatively competitive IoU of 23.9 on S2Looking. However, its performance on LEVIR-CD and WHU-CD is much lower than ours. Similarly, the APE - / - DINOv2 configuration reaches a IoU of 55.2 on WHU-CD, but its result on S2Looking dataset drops sharply to 5.3. In contrast, OmniOVCD achieves consistently high and stable performance, with IoU scores of 67.2, 66.5, and 24.5 on LEVIR-CD, WHU-CD, and S2Looking, respectively. This demonstrates that our method performs reliably across different building shapes and complex environments.

\textbf{Land Cover Change Detection.} Results on the SECOND dataset are shown in Tab.~\ref{tab:second}. We re-implemented the four best configurations from DynamicEarth~\protect\cite{li2025dynamicearth}. As described in the table, OmniOVCD clearly outperforms them on all categories. Our method achieves higher scores not only in class average IoU and F1, but also for every individual category. For example, compared to the strongest baseline, OmniOVCD improves the IoU for the ``Building'' and ``Water'' classes from 38.8 and 15.3 to 45.2 and 21.2, respectively. This demonstrates that our framework achieves strong and consistent performance on the OVCD task.

\textbf{Efficiency Comparison.} As shown in Fig.~\ref{fig:cost}, OmniOVCD not only achieves the highest performance, but also exhibits the fastest inference speed and the lowest memory usage, further demonstrating the superiority of our architecture.

\subsection{Ablation Study}
\textbf{Effects of Semantic Head Fusion.} This ablation study examines whether fusing semantic head outputs is necessary in the SFID strategy before instance decoupling. Although the framework can generate instance-level masks directly, fusing semantic head outputs provides more stable guidance for subsequent stages. We verify this by comparing the full OmniOVCD with a variant that skips semantic head fusion.

As shown in Tab.~\ref{tab:abs-fusion}, removing semantic head fusion results in clear performance drops. This indicates that the semantic head provides important category-level guidance. By fusing semantic head outputs, the SFID strategy produces a more accurate and informative semantic representation, leading to improved instance-level change predictions.

\begin{table}[t]
\centering
\begin{tabular}{c|cc|cc}
\toprule
\multirow{2}{*}{Fusion Strategy}                      & \multicolumn{2}{c|}{LEVIR-CD}     & \multicolumn{2}{c}{WHU-CD} \\ \cmidrule(l){2-5} 
                                       & IoU  & \multicolumn{1}{c|}{F1}   & IoU          & F1          \\ \midrule
\multicolumn{1}{c|}{w/o Semantic Head} & 53.0 & \multicolumn{1}{c|}{69.3} & 52.9         & 64.5        \\ \rowcolor{blue!5}
w/ Semantic Head                       & \textbf{67.2} & \textbf{80.4}                      & \textbf{66.5}         & \textbf{79.9}        \\ \bottomrule
\end{tabular}
\caption{Effects of semantic head fusion.}
\label{tab:abs-fusion}
\end{table}

\begin{table}[t]
\centering
\begin{tabular}{c|cc|cc}
\toprule
\multirow{2}{*}{Matching Strategy} & \multicolumn{2}{c}{LEVIR-CD}  & \multicolumn{2}{c}{WHU-CD}    \\ \cmidrule(l){2-5} 
            & IoU  & F1   & IoU  & F1   \\ \cmidrule(r){1-5}
PMC         & 29.2 & 45.2 & 26.3 & 41.7 \\
L1 Distance & 60.6 & 75.5 & 56.9 & 72.6 \\
L2 Distance & 59.1 & 74.3 & 47.6 & 64.5 \\ \rowcolor{blue!5}
Instance Matching                  & \textbf{67.2} & \textbf{80.4} & \textbf{66.5} & \textbf{79.9} \\ \bottomrule
\end{tabular}
\caption{Effects of change mask match strategy.}
\label{tab:match_strategy}
\end{table}

\textbf{Effects of Change Mask Match Strategy.} To evaluate the instance decoupling and matching strategy, we compare it with three other change detection strategies. The simplest is Pixel-wise Mask Comparison (PMC), which detects changes by comparing corresponding pixels in the bi-temporal semantic masks. We also implement two logits-based methods: Logit-space $L_1$ Distance and Logit-space $L_2$ Distance. These methods compute the $L_1$ (sum of absolute differences) or $L_2$ (Euclidean) distance between the bi-temporal logits at each pixel, followed by a thresholding step to generate change masks. 

As shown in Tab.~\ref{tab:match_strategy}, our instance matching strategy consistently achieves the best results. The simpler pixel-wise methods, such as PMC and logit-based $L_1$ or $L_2$ distances, often struggle with noise at the pixel level. PMC can produce fragmented change predictions when there are isolated classification errors, and the $L_1$ and $L_2$ distances are sensitive to small differences in image intensity or lighting between the two time points. In contrast, our instance matching strategy operates at the instance level. It compares decoupled instance masks across time, which reduces the influence of pixel-level noise and helps preserve the precise shape and boundaries of each changed instance. These results show that the instance-level comparison is crucial for accurate change detection.

\begin{figure}[t]
  \centering
   \includegraphics[width=0.95\linewidth]{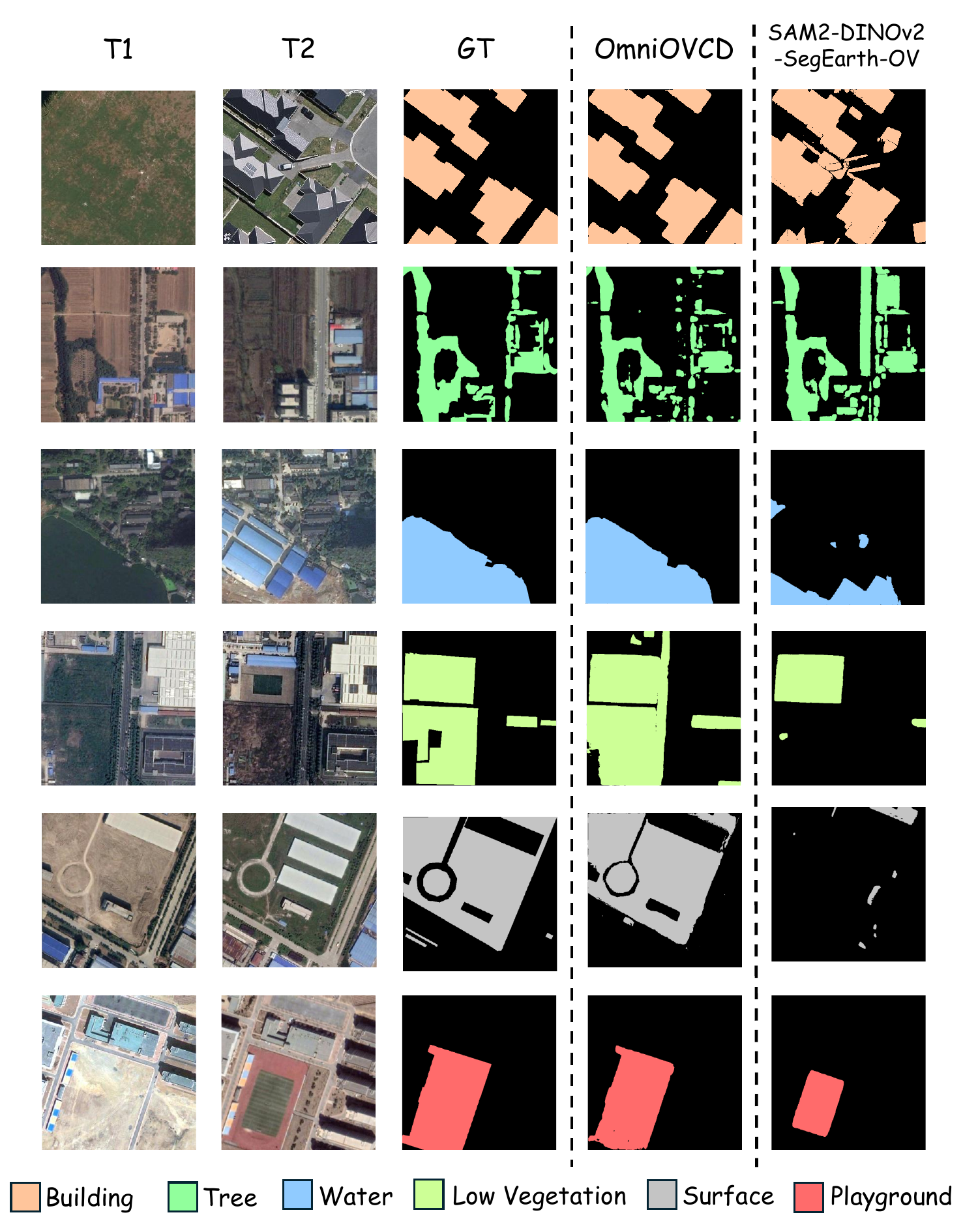}
   \caption{Visual comparisons of OmniOVCD with other state-of-the-art methods for open-vocabulary change detection.}
   \label{fig:vis}
\end{figure}

\subsection{Visualization Results}
We qualitatively compare OmniOVCD with the best-performing configuration from DynamicEarth~\cite{li2025dynamicearth}, namely SAM2-DINOv2-SegEarth-OV. As shown in Fig.~\ref{fig:vis}, OmniOVCD delivers consistently accurate change detection across all categories, while the DynamicEarth configuration exhibits false negatives and false positives in complex scenes. These results further validate the robustness and effectiveness of OmniOVCD.

\section{Conclusion}
In this paper, we propose OmniOVCD, the first standalone framework designed to streamline the open-vocabulary change detection task. The framework employs the Synergistic Fusion to Instance Decoupling (SFID) strategy to fully leverage the unified architectural strengths of SAM 3. This design allows OmniOVCD to address the instability often found in traditional pipelined methods. Extensive evaluations across four major benchmarks show that OmniOVCD not only achieves state-of-the-art performance but also maintains strong stability and generalization in complex open-world scenarios. We believe this framework establishes a robust and efficient foundation for future advancements in intelligent remote sensing analysis.

\bibliographystyle{named}
\bibliography{ijcai26}

\end{document}